\PassOptionsToPackage{table}{xcolor}
\documentclass[11pt]{article}

\usepackage[preprint]{acl}

\usepackage{times}
\usepackage{latexsym}
\usepackage{fancyvrb}
\usepackage[T1]{fontenc}

\usepackage[utf8]{inputenc}

\usepackage{microtype}
\usepackage{amssymb}
\usepackage{inconsolata}

\usepackage{CJKutf8}
\usepackage{graphicx}
\usepackage{enumitem}
\usepackage{amsmath}
\usepackage{booktabs}
\usepackage{multirow}
\usepackage{mdframed}
\usepackage{tcolorbox}
\usepackage{hyperref}
\usepackage{marvosym}

\tcbuselibrary{skins,breakable}

\usepackage{booktabs}
\usepackage{multirow}
\usepackage{graphicx}
\usepackage{colortbl}

\newif\ifshowtodo
\showtodotrue   

\ifshowtodo
  \newcommand{\TODO}[1]{%
    \textcolor{red}{\textbf{TODO:} #1}%
  }
  \newcommand{\CTODO}[1]{%
    \begin{CJK}{UTF8}{gbsn}%
    \TODO{#1}%
    \end{CJK}%
  }
\else
  \newcommand{\TODO}[1]{}
  \newcommand{\CTODO}[1]{}
\fi

\definecolor{mylightblue}{RGB}{100,149,237}
\tcbset{
  enhanced, 
  colback=white!200!black, 
  colframe=mylightblue, 
  colbacktitle=mylightblue, 
  title filled, 
  coltitle=white, 
  fonttitle=\bfseries, 
  arc=3mm, 
  outer arc=3mm, 
  boxrule=0.5mm, 
  toprule=0.5mm, 
  bottomrule=0.5mm, 
  titlerule=0.5mm, 
  drop fuzzy shadow, 
}
\setlist[itemize]{topsep=1pt, itemsep=0.5pt}

\usepackage[table]{xcolor} 
\definecolor{ourblue}{HTML}{F0F5FF} 
\usepackage{tabularx}

\title{MobEvolve: An Agentic Self-Evolving Heuristic System for Interpretable Human Mobility Generation}

\newcommand{\equalcontrib}{\textsuperscript{$\heartsuit$}}
\newcommand{\corresponding}{\textsuperscript{\Letter}}

\author{
  \textbf{Junlin He\textsuperscript{1}\equalcontrib},
  \textbf{Yihong Tang\textsuperscript{2,8}\equalcontrib},
  \textbf{Tong Nie\textsuperscript{1}},
  \textbf{Ao Qu\textsuperscript{4}},
  \textbf{Yuebing Liang\textsuperscript{5}}, \\
  \textbf{Hamzeh Alizadeh\textsuperscript{6}},
  \textbf{Bang Liu\textsuperscript{7,8}},
  \textbf{Wei Ma\textsuperscript{1}\corresponding},
  \textbf{Lijun Sun\textsuperscript{2}\corresponding}
  \\[0.75em]
  \textsuperscript{1}The Hong Kong Polytechnic University,\,
  \textsuperscript{2}McGill University,\,
  \textsuperscript{4}MIT, \,
  \textsuperscript{5}Tsinghua University,\, \\
  \textsuperscript{6}Autorité régionale de transport métropolitain, 
  \textsuperscript{7}Université de Montréal, \,\\
  \textsuperscript{8}Mila -- Quebec AI Institute \\
    [0.5mm]
    \tt\small \href{mailto:junlinspeed.he@connect.polyu.hk}{junlinspeed.he@connect.polyu.hk} \, \href{mailto:yihong.tang@mail.mcgill.ca}{yihong.tang@mail.mcgill.ca} \,
  \href{mailto:wei.w.ma@polyu.edu.hk}{wei.w.ma@polyu.edu.hk}  \,
  \href{mailto:lijun.sun@mcgill.ca}{lijun.sun@mcgill.ca}
}

\begin{document}
\maketitle

\begin{abstract}
Human mobility generation aims to synthesize realistic trip chains for target populations based on individual features. Existing paradigms, including deep generative models, LLM-based methods, and traditional heuristics, struggle to satisfy the complex demands of this task while simultaneously maintaining interpretability, behavioral plausibility, population-level distributional alignment, and inference efficiency. To bridge this gap, we introduce \textbf{MobEvolve}, the first agentic self-evolving heuristic framework for human mobility generation. 
MobEvolve initializes a behavior-inspired heuristic system and employs an LLM agent to iteratively evolve its internal logic. By diagnosing empirical misalignments and failure cases on a validation set, the agent proposes targeted updates and accumulates evolution memory for cumulative self-improvement. Extensive evaluations on the Singapore and Montreal benchmarks demonstrate that MobEvolve significantly outperforms state-of-the-art deep generative and LLM-based methods in individual trajectory fidelity, population-level distribution alignment, and behavioral plausibility, while preserving interpretability and high inference efficiency.
\end{abstract}

\begingroup
\renewcommand\thefootnote{}\footnotetext{
$^\heartsuit$ Equal contribution.
\textsuperscript{\Letter}\ Corresponding authors.
}
\endgroup

\section{Introduction}

\begin{figure}[t]
    \centering
    \includegraphics[width=\linewidth]{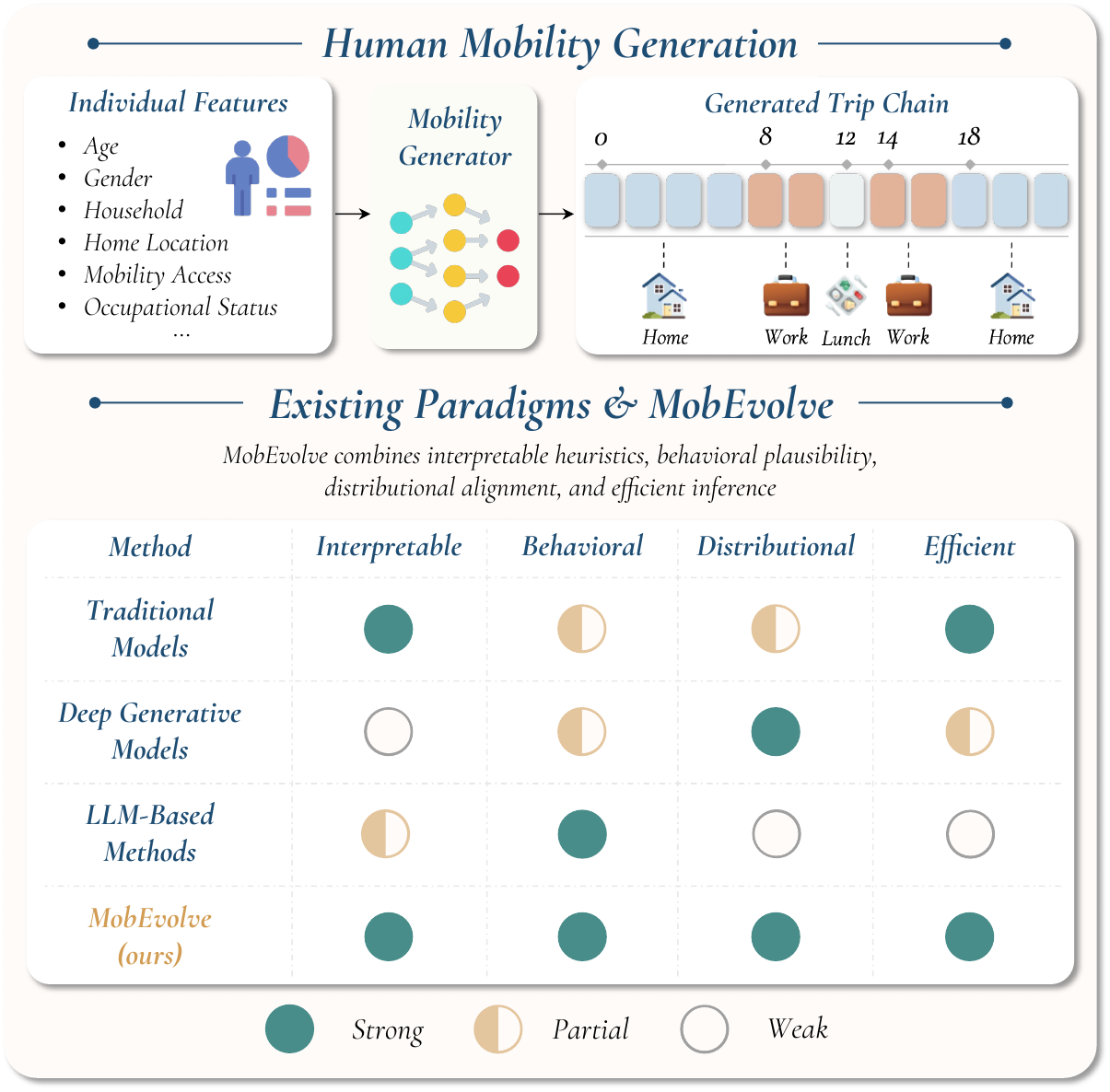}
    \vspace{-17pt}
    \caption{Overview of MobEvolve. Top: feature-conditioned synthesis of typical-day trip chains. Bottom: MobEvolve brings together interpretability, behavioral plausibility, distributional alignment, and efficiency.}
    \vspace{-14pt}
\label{fig:intro}
\end{figure}

Urban planners demand realistic daily trip chains for scenario evaluation and evidence-based decision support~\cite{beckman1996creating,w2016multi,wang2025developing}. While empirical mobility observations are often sparse and privacy-constrained, individual features (e.g., demographics, household structures, and mobility access) are more readily available~\cite{de2013unique,amichi2025exploring,bittencourt2026role}. This creates a practical need: synthesizing mobility trip chains from \textit{individual features}. Unlike traditional mobility prediction that relies on historical trajectories~\cite{wu2021individual,zhang2018mobility,long2025universal}, our task focuses on a \emph{feature-conditioned synthesis} problem, often operating under a regime where only limited paired observations are available for training or retrieval~\cite{wang2024large,russo2025framework,liang2026analyzing}. 

As shown in Figure~\ref{fig:intro}, this task presents a feature-conditioned
synthesis challenge with two core requirements. First, \textbf{Behavioral
Plausibility}: each synthesized trip chain should be consistent with the
individual's socio-demographic constraints and basic activity-travel logic~\cite{kitamura1997generation,rezvany2024review}.
Second, \textbf{Distributional Alignment}: the generated population should
statistically match the aggregate mobility patterns of the real population~\cite{gonzalez2008understanding,song2010limits}.
For large-scale policy analysis, the synthesis framework should also be an
\textbf{interpretable white-box} with \textbf{computational efficiency}, so that
generated trip chains can be inspected and scaled to population-level
simulation~\cite{adenaw2022generating,nguyen2025large}.

Existing mobility generation paradigms satisfy only part of these requirements. Deep generative models and LLM-based approaches have recently become prominent directions~\cite{luca2021survey,guo2025language,da2025generative}. Deep generative models~\cite{kingma2013auto,goodfellow2020generative,liang2026analyzing} can learn complex mobility distributions, but their black-box generation mechanisms are difficult to inspect and, without explicit behavioral constraints, may produce implausible trip chains, such as a student attending school on a weekend night. LLM-based methods~\cite{tang2024itinera,wang2024large} can leverage semantic knowledge to generate plausible trajectories, but individual-level generation is costly and difficult to align with population distributions. Traditional heuristic systems~\cite{bowman2001activity,arentze2004learning} are efficient and interpretable, but depend on manually designed rules whose
incomplete coverage can degrade generation quality.

To bridge this gap, we draw inspiration from recent advancements in LLM coding agents~\cite{openai_codex_2026,anthropic_claude_code_2026}, which demonstrate the capability to inspect executable systems, interpret evaluation feedback, iteratively revise internal logic, and utilize prior trials as context~\cite{qu2026coral,weng2026learning_beyond_gradients}. This paradigm shift presents a new opportunity to develop \textit{self-evolving} mobility heuristics. By empowering an initial heuristic system to automatically discover and iteratively refine itself from complex data, we eliminate the bottleneck of manual rule and complex system design. Consequently, this evolving system achieves a favorable combination of these aspects: it preserves the \emph{interpretability}, \emph{behavioral plausibility}, and \emph{scalable inference}, while achieving \emph{distribution alignment} through self-evolution.

In this work, we introduce \textbf{MobEvolve}, an agentic self-evolving heuristic system for interpretable human mobility generation. MobEvolve initializes an interpretable mobility heuristic system $G^{(0)}$ and employs an LLM agent to iteratively evolve its internal generation logic. Specifically, we partition the empirical observations of a city into training, validation, and test sets, where each individual is represented by a single trip chain. The initial heuristic system leverages the training set to construct a template library and compute empirical mobility distributions. During the evolution phase, the agent evaluates the system's performance on the validation set, analyzes quantitative metrics alongside specific failure cases, and proposes targeted heuristic updates. By accumulating attempted trials in an evolution memory, the agent effectively translates population-level generation errors into auditable heuristic refinements. The evolved system is finally evaluated on the \emph{unseen} test set. The contributions are summarized as follows:
\begin{itemize}[leftmargin=*]
    \item We introduce \textbf{MobEvolve}, the first agentic self-evolving heuristic framework for human mobility generation. It successfully unifies interpretable heuristic modeling, behaviorally plausible trip-chain synthesis, empirical distributional alignment, and efficient population-scale generation.
    \item We design a novel data-driven evolution mechanism. MobEvolve initializes its heuristics using training-set template libraries and evolves them through a validation-driven attribution process. By iteratively diagnosing empirical misalignments and failure cases, applying targeted updates, and preserving evolution memory, the system achieves cumulative self-improvement.
    \item We conduct extensive evaluations on the Singapore and Montreal benchmarks. MobEvolve significantly outperforms state-of-the-art deep generative and LLM-based methods across three critical dimensions: individual-level trip fidelity, population-level distribution alignment, and behavioral plausibility. Crucially, it achieves these superior results while strictly preserving the inherent interpretability and high inference efficiency of heuristic systems.
\end{itemize}

\section{Related Work}
\label{sec:related-work}

\paragraph{Human Mobility Generation}
Human mobility generation aims to synthesize realistic trip chains that reproduce both individual travel behavior and population-level mobility distributions~\cite{brockmann2006scaling,barbosa2018human}. Early models explicitly encode activity participation, scheduling, and spatial choice, making them interpretable and planning-compatible, but often manually calibrated~\cite{bowman2001activity,arentze2004learning}. Routine-based generators further show that empirical mobility can be synthesized by separating temporal routines from spatial movement~\cite{simini2012universal,pappalardo2018data}. More recent deep generative models learn complex mobility patterns directly from data~\cite{kapp2023generative,luca2021survey,liang2026analyzing}, improving distributional flexibility but often reducing interpretability and explicit constraint control.  LLM-based systems such as ItiNera, LLMob, and MobAgent use semantic reasoning, retrieval, and individual profiles for personalized mobility generation~\cite{wang2024large}. LLMSynthor further improves LLM-based synthesis with proposal sampling for efficiency and iterative feedback for distributional alignment~\cite{tang2025llmsynthor}. However, LLM-based synthesis remains less transparent and may inherit LLM biases. Overall, prior work improves different dimensions of mobility generation, but does not jointly provide interpretability, behavioral plausibility, distributional alignment, and efficient generation.

\vspace{-3pt}
\paragraph{Self-Evolving Agents}
Recent work shows that LLM agents can improve executable systems through iterative proposal, evaluation, and accumulated experience. FunSearch applies LLM-guided program search to mathematical discovery~\cite{romera2024mathematical}, Eureka evolves reward code from task feedback~\cite{ma2024eureka}, and AlphaEvolve extends coding-agent evolution to scientific and algorithmic discovery~\cite{novikov2025alphaevolve}. More recent systems emphasize self-improvement over time: the Darwin G\"odel Machine modifies its own coding-agent code and validates changes on benchmarks~\cite{zhang2025darwin}; CORAL studies autonomous multi-agent evolution with persistent memory and evaluator separation~\cite{qu2026coral}; and Meta-Harness optimizes LLM harness code using prior candidates, scores, and execution traces~\cite{lee2026meta}. These works establish self-evolution as a mechanism for improving programs, agents, and harnesses without gradient updates. MobEvolve extends this idea to structured human mobility generation problem, where the agent improves mobility heuristics tied to generation decisions. This yields a fully interpretable generator whose evolution is traceable from empirical misalignment to heuristic update.

\begin{figure*}[t]
    \centering
    \includegraphics[width=\linewidth]{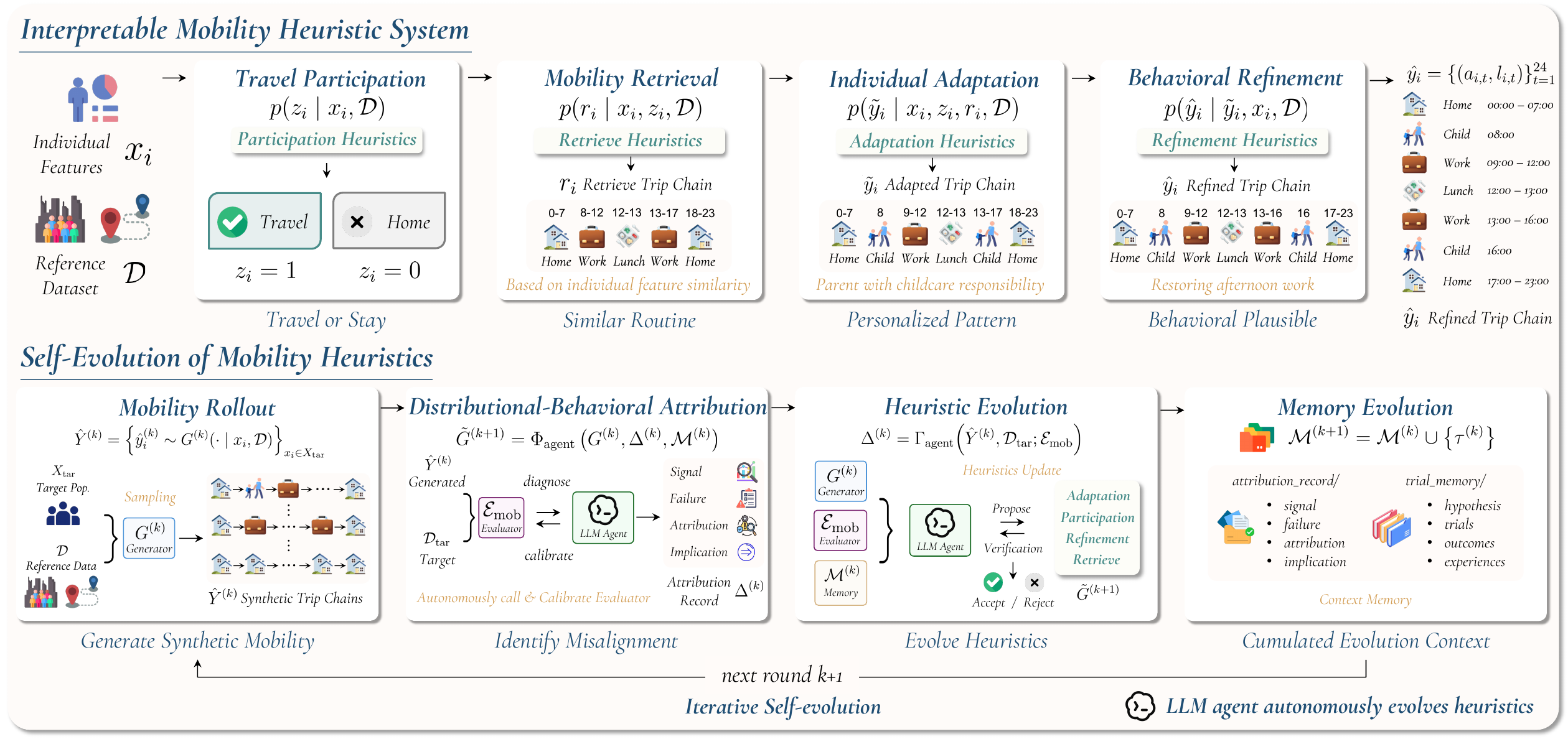}
    \vspace{-21pt}
    \caption{Overview of MobEvolve. The upper panel illustrates the interpretable mobility heuristic system, which factorizes trip-chain generation into travel participation, mobility retrieval, mobility adaptation, and behavioral refinement. The lower panel shows the self-evolution process: the current generator is rolled out on the target population, distributional-behavioral attribution identifies empirical misalignments, the LLM agent proposes and verifies heuristic updates, and evolution memory accumulates attribution records and trials for subsequent evolutions.}
    \vspace{-12pt}

\label{fig:mobevolve}
\end{figure*}

\vspace{-2pt}
\section{Problem Formulation}
\vspace{-2pt}

Let $x_i\in\mathcal{X}$ denote the individual features (e.g., demographics) of person $i$, and let $y_i=\{(a_{i,t},l_{i,t})\}_{t=1}^{T}$ denote a typical-day mobility trip chain over $T=24$ hourly slots, and $a_{i,t}\in\mathcal{A}$ and $l_{i,t}\in\mathcal{L}$ denote the activity and location states at time slot $t$. Given training set of empirical observations as reference mobility data $\mathcal{D}=\{(x_j,y_j)\}_{j=1}^{M}$, the task is to generate trip chains for a target population $X_{\mathrm{tar}}=\{x_i\}_{i=1}^{N}$ of \emph{unseen} individuals. Let $\mathcal{G}$ denote the space of conditional mobility generators. We seek a generator $G\in\mathcal{G}$ that produces realistic trip chains aligned with target mobility behavior:
\begin{equation}
\label{eq:prob_form}
\begin{aligned}
G^{\star}=
&\arg\min_{G\in\mathcal{G}} \;
 \mathcal{E}_{\mathrm{mob}}\!\left(
\{\hat{y}_i\}_{i=1}^{N},\,\mathcal{D}_{\mathrm{tar}}
\right), \\
&\text{s.t.}\quad
 \hat{y}_i\sim G(\cdot\mid x_i,\mathcal{D}),
\quad \forall x_i\in X_{\mathrm{tar}}.
\end{aligned}
\end{equation}
$\mathcal{D}_{\mathrm{tar}}=\{(x_i,y_i)\}_{i=1}^{N}$ is the target observations used for evaluation, and $\mathcal{E}_{\mathrm{mob}}$ is a evaluator of behavioral plausibility and distributional alignment.

\vspace{-2pt}
\section{Methodology}
\vspace{-2pt}

Figure~\ref{fig:mobevolve} provides an overview of MobEvolve. The method has two parts: an interpretable mobility heuristic system that generates trip chains from individual features and reference data, and a self-evolution process that improves the internal heuristic logic through evaluator-guided agent updates.

\vspace{-2pt}
\subsection{Interpretable Mobility Heuristic System}

Because human mobility reflects activity needs, habitual routines, spatial opportunities, and validity constraints, a generator should make these behavioral assumptions \emph{explicit} while matching empirical distributions. We therefore initialize $G^{(0)}$ with an interpretable and behaviorally grounded heuristic system as the evolvable basis. Given $x_i$, MobEvolve factorizes the generation of $\hat{y}_i$ as:
\begin{equation}
\begin{aligned}
p(\hat{y}_i \mid x_i)=\,&p(z_i \mid x_i) \,
   p(r_i \mid x_i,z_i) \\
 \quad \quad \cdot \,&p(\tilde{y}_i \mid x_i,z_i,r_i) \,
   p(\hat{y}_i \mid \tilde{y}_i,x_i),
\end{aligned}
\label{eq:mobility_heuristic_factorization}
\end{equation}
where all components are conditioned on $\mathcal{D}$, which is omitted for simplicity. Here, $z_i$ denotes travel participation, $r_i$ denotes the retrieved reference trip chain, $\tilde{y}_i$ denotes its adapted form, and $\hat{y}_i$ is the output after behavioral refinement.

\vspace{-5pt}
\paragraph{Travel Participation.}
Activity-based travel theory treats travel as derived from activity participation and schedule formation~\cite{bowman2001activity,arentze2000albatross}. This motivates a participation heuristic: $p(z_i \mid x_i, \mathcal{D})$ determines whether person $i$ conducts out-of-home activities given individual features $x_i$.

\vspace{-5pt}
\paragraph{Mobility Retrieval.}
Human mobility is highly regular, predictable, and routine-based~\cite{gonzalez2008understanding,song2010limits,eagle2009eigenbehaviors,pappalardo2018data}. This motivates a retrieval heuristic: $p(r_i \mid x_i,z_i,\mathcal{D})$ selects an empirical trip chain from behaviorally similar records as the reference pattern for generation.

\vspace{-5pt}
\paragraph{Individual Adaptation.}
Routine-based behavior varies across individuals~\cite{gonzalez2008understanding,song2010limits}. This motivates an adaptation heuristic: $p(\tilde{y}_i \mid x_i,z_i,r_i,\mathcal{D})$ adjusts the retrieved pattern to reflect individual behavioral variation.

\vspace{-5pt}
\paragraph{Behavioral Refinement.}
Mobility generation requires behavioral plausibility and internal consistency~\cite{tang2025llmsynthor}. This motivates a refinement heuristic: $p(\hat{y}_i \mid \tilde{y}_i,x_i, \mathcal{D})$ transforms the adapted trip chain into a coherent output with realistic daily mobility structure.

This factorization makes $G^{(0)}$ an evolvable heuristic basis: the four behavioral decisions remain for behavioral interpretability, while their internal logic can be evolved through self-evolution.

\subsection{Self-Evolution of Mobility Heuristics}

Starting from $G^{(0)}$, MobEvolve treats the internal generation logic of the four components as evolvable mobility heuristics. In each round, an LLM agent autonomously evolves these heuristics through four stages: \emph{Mobility Rollout}, which applies the current generator to the target population features $X_{\mathrm{tar}}$ using reference data $\mathcal{D}$; 
\emph{Distributional-Behavioral Attribution}, which identifies distributional and behavioral misalignments by comparing generated trip chains with $\mathcal{D}_{\mathrm{tar}}$; \emph{Heuristic Evolution}, which proposes and verifies a targeted heuristic update; and \emph{Memory Evolution}, which records each trial and consolidates useful heuristic evidence for future rounds.

\vspace{-5pt}
\paragraph{Mobility Rollout.}
At round $k$, the generator $G^{(k)}$ is rolled out on the evolution population features $X_{\mathrm{tar}}$, which produces synthetic trip chains:
\begin{equation}
\hat{Y}^{(k)}
=
\left\{
\hat{y}^{(k)}_i \sim G^{(k)}(\cdot \mid x_i,\mathcal{D})
\right\}_{x_i\in X_{\mathrm{tar}}}.
\end{equation}
This stage exposes the behavioral consequences of the current heuristics at population scale.

\vspace{-5pt}
\paragraph{Distributional-Behavioral Attribution.}
MobEvolve compares $\hat{Y}^{(k)}$ with the empirical trip chains in $\mathcal{D}_{\mathrm{tar}}$ using the evaluator $\mathcal{E}_{\mathrm{mob}}$. The evaluation is compiled into an attribution record $\Delta^{(k)}$:
\begin{equation}
\Delta^{(k)}
=
\Gamma_{\mathrm{agent}}\!\left(
\hat{Y}^{(k)}, \mathcal{D}_{\mathrm{tar}};
\mathcal{E}_{\mathrm{mob}}
\right),
\end{equation}
where $\Gamma_{\mathrm{agent}}$ denotes the agent's attribution process over evaluator outputs and its autonomous checks. The resulting $\Delta^{(k)}$ summarizes the agent's interpretation of how the current mobility heuristics deviate from empirical behavior.
This stage helps the agent identify the relevant behavioral decision and formulate a direction for future evolution.

\vspace{-5pt}
\paragraph{Heuristic Evolution.}
Given $G^{(k)}$, the attribution record $\Delta^{(k)}$, and the current evolution memory $\mathcal{M}^{(k)}$, the LLM agent proposes a candidate generator by updating the relevant mobility heuristics:
\begin{equation}
\tilde{G}^{(k+1)}
=
\Phi_{\mathrm{agent}}
\left(
G^{(k)}, \Delta^{(k)}, \mathcal{M}^{(k)}
\right).
\end{equation}
The update is targeted to the internal generation logic of travel participation, mobility retrieval, mobility adaptation, or behavioral refinement. The behavioral structure of $G^{(0)}$ gives the agent a meaningful search space: the agent can relate observed misalignments to heuristics and revise them accordingly. The agent autonomously proposes the heuristic revision and verifies the candidate under $\mathcal{E}_{\mathrm{mob}}$. The candidate is promoted to $G^{(k+1)}$ only when empirical verification supports it as an acceptable update. Otherwise, the current generator is carried forward, while the candidate is retained in memory for future evolution.

\vspace{-5pt}
\paragraph{Memory Evolution.}
After verification, MobEvolve updates the evolution memory with the completed heuristic evolution trial:
\begin{equation}
\mathcal{M}^{(k+1)}
=
\mathcal{M}^{(k)}
\cup
\left\{
\tau^{(k)}
\right\}.
\end{equation}
Here, $\tau^{(k)}$ records the attribution record, the proposed heuristic update, the agent rationale, the verification outcome, and whether the candidate is promoted to $G^{(k+1)}$ or rejected. The memory provides evolving context for later rounds through two records: \texttt{attribution\_record/}, which links current evaluation signals to localized behavioral misalignments and representative generated trip chains; and \texttt{trial\_memory/}, which stores prior interventions, hypotheses, outcomes, and experiences from accepted and rejected trials. Together, these records make self-evolution cumulative: promoted trials form an evolving trajectory of effective mobility heuristics, while rejected trials preserve negative heuristic evidence. This allows later rounds to connect present failures with prior evidence, and provides an auditable record of how the mobility generator evolves over time. Appendix~\ref{appx:implementation} provides implementation details of the evolution context.

\begin{table*}[t]
\centering
\scriptsize
\definecolor{ourblue}{HTML}{F0F5FF}
\setlength{\tabcolsep}{3pt}
\renewcommand{\arraystretch}{1}
\caption{Held-out test results on representative metrics. Lower values indicate better performance ($\downarrow$). Individual metrics assess hourly time-slot recovery, using joint activity-location labels for Montreal. Distribution metrics measure population-level calibration, while Behavior metrics assess task-specific plausibility and consistency.}
\vspace{-5pt}
\label{tab:main-results}
\resizebox{\linewidth}{!}{%
\begin{tabular}{cl ccc ccc ccc c}
\toprule
& \multirow{2}{*}{\textbf{Method}} & \multicolumn{3}{c}{\textbf{Individual} $\downarrow$} & \multicolumn{3}{c}{\textbf{Distribution} $\downarrow$} & \multicolumn{3}{c}{\textbf{Behavior} $\downarrow$} & \textbf{Overall} $\downarrow$ \\
\cmidrule(lr){3-5}\cmidrule(lr){6-8}\cmidrule(lr){9-11}\cmidrule(lr){12-12}
& & Score & Slot Err. & F1 Err. & Score & Slot JSD & Seg. JSD & Score & Night & Travel & Score \\
\midrule
\multirow{14}{*}{\rotatebox[origin=c]{90}{\textbf{Singapore}}}
& \multicolumn{11}{>{\columncolor[HTML]{F9F9F9}}l}{\emph{\textbf{$\blacktriangledown$ Traditional Mobility Models}}} \\
& MNL                        & 0.4737 & 0.2476 & 0.2261 & 0.2316 & 0.0137 & 0.0636 & 0.9827 & 0.0710 & 0.0632 & 1.6880 \\
& Markov                     & 0.7587 & 0.3736 & 0.3851 & 0.4791 & 0.0062 & 0.2171 & 1.9065 & 0.0548 & 0.0180 & 3.1443 \\
& Gaussian Copula            & 0.6353 & 0.3182 & 0.3171 & 0.4520 & 0.0014 & 0.1539 & 2.3198 & 0.0290 & 0.0592 & 3.4071 \\
\cmidrule(lr){2-12}
& \multicolumn{11}{>{\columncolor[HTML]{F9F9F9}}l}{\emph{\textbf{$\blacktriangledown$ Deep Generative Models}}} \\
& Deep Behavior Choice       & 0.4131 & 0.2117 & 0.2014 & 0.1169 & 0.0066 & 0.0357 & 0.2683 & 0.0048 & 0.0105 & 0.7983 \\
& AIRL Behavior Choice       & 0.3876 & 0.2006 & 0.1870 & 0.1098 & 0.0061 & 0.0335 & 0.2880 & 0.0024 & 0.0208 & 0.7854 \\
& TVAE                       & 0.7668 & 0.3834 & 0.3834 & 0.4784 & 0.0036 & 0.2128 & 1.9014 & 0.0876 & 0.0152 & 3.1466 \\
& CopulaGAN                  & 0.7927 & 0.4021 & 0.3906 & 0.5836 & 0.0095 & 0.2026 & 2.4513 & 0.1217 & 0.0484 & 3.8276 \\
& CTGAN                      & 0.8221 & 0.4200 & 0.4021 & 0.7563 & 0.0156 & 0.2828 & 2.5986 & 0.1528 & 0.0563 & 4.1770 \\
\cmidrule(lr){2-12}
& \multicolumn{11}{>{\columncolor[HTML]{F9F9F9}}l}{\emph{\textbf{$\blacktriangledown$ LLM-based Mobility Generation}}} \\
& LLMob Few-shot             & 0.4949 & 0.2587 & 0.2362 & 0.4564 & 0.0204 & 0.1496 & 0.5445 & 0.0274 & 0.0104 & 1.4958 \\
& LLMob Zero-shot            & 0.7731 & 0.4173 & 0.3558 & 2.3958 & 0.1785 & 0.5079 & 5.9870 & 0.4172 & 0.0794 & 9.1559 \\
\cmidrule(lr){2-12}
& \cellcolor{ourblue}\textbf{MobEvolve (ours)}
& \cellcolor{ourblue}\textbf{0.3839}
& \cellcolor{ourblue}\textbf{0.1920}
& \cellcolor{ourblue}\textbf{0.1919}
& \cellcolor{ourblue}\textbf{0.0578}
& \cellcolor{ourblue}\textbf{0.0005}
& \cellcolor{ourblue}\textbf{0.0272}
& \cellcolor{ourblue}\textbf{0.0759}
& \cellcolor{ourblue}\textbf{0.0014}
& \cellcolor{ourblue}\textbf{0.0066}
& \cellcolor{ourblue}\textbf{0.5176} \\

\midrule

\multirow{14}{*}{\rotatebox[origin=c]{90}{\textbf{Montreal}}}
& \multicolumn{11}{>{\columncolor[HTML]{F9F9F9}}l}{\emph{\textbf{$\blacktriangledown$ Traditional Mobility Models}}} \\
& MNL                        & 1.6181 & 0.4183 & 0.3984 & 0.2369 & 0.0092 & 0.0484 & 0.6267 & 0.0315 & 0.0660 & 2.4817 \\
& Markov                     & 1.7048 & 0.4229 & 0.4283 & 0.4474 & 0.0018 & 0.1644 & 0.8258 & 0.0066 & 0.1196 & 2.9780 \\
& Gaussian Copula            & 1.6806 & 0.4206 & 0.4184 & 0.2615 & 0.0017 & 0.1095 & 1.5524 & 0.2703 & 0.0947 & 3.4945 \\
\cmidrule(lr){2-12}
& \multicolumn{11}{>{\columncolor[HTML]{F9F9F9}}l}{\emph{\textbf{$\blacktriangledown$ Deep Generative Models}}} \\
& Deep Behavior Choice       & 1.5297 & 0.3832 & 0.3730 & 0.1488 & 0.0046 & 0.0496 & 0.2494 & 0.0146 & 0.0212 & 1.9279 \\
& AIRL Behavior Choice       & 1.5048 & 0.3798 & 0.3729 & 0.1186 & 0.0023 & 0.0463 & 0.2556 & 0.0069 & 0.0176 & 1.8790 \\
& TVAE                       & 1.6672 & 0.4172 & 0.4193 & 0.3719 & 0.0048 & 0.1469 & 0.5715 & 0.0531 & 0.0698 & 2.6106 \\
& CopulaGAN                  & 1.7984 & 0.4492 & 0.4383 & 0.4707 & 0.0104 & 0.1792 & 1.3733 & 0.1581 & 0.0962 & 3.6424 \\
& CTGAN                      & 1.8256 & 0.4632 & 0.4408 & 0.5425 & 0.0119 & 0.2106 & 1.5133 & 0.1762 & 0.0996 & 3.8814 \\
\cmidrule(lr){2-12}
& \multicolumn{11}{>{\columncolor[HTML]{F9F9F9}}l}{\emph{\textbf{$\blacktriangledown$ LLM-based Mobility Generation}}} \\
& LLMob Few-shot             & 2.0394 & 0.5109 & 0.4523 & 1.1456 & 0.0531 & 0.3944 & 0.6128 & 0.0672 & 0.0187 & 3.7978 \\
& LLMob Zero-shot            & 3.1245 & 0.8168 & 0.7385 & 4.7400 & 0.5280 & 0.7526 & 3.9169 & 0.5102 & 0.7025 & 11.7814 \\
\cmidrule(lr){2-12}
& \cellcolor{ourblue}\textbf{MobEvolve (ours)}
& \cellcolor{ourblue}\textbf{1.4530}
& \cellcolor{ourblue}\textbf{0.3678}
& \cellcolor{ourblue}\textbf{0.3706}
& \cellcolor{ourblue}\textbf{0.1060}
& \cellcolor{ourblue}\textbf{0.0009}
& \cellcolor{ourblue}\textbf{0.0425}
& \cellcolor{ourblue}\textbf{0.1172}
& \cellcolor{ourblue}\textbf{0.0039}
& \cellcolor{ourblue}\textbf{0.0149}
& \cellcolor{ourblue}\textbf{1.6762} \\
\bottomrule
\end{tabular}
}
\vspace{-8pt}
\end{table*}

\vspace{-2pt}
\section{Experiments}
\vspace{-2pt}

\label{sec:experiments}

We evaluate MobEvolve from three perspectives: generation quality, self-evolution dynamics, and interpretability. We first describe the datasets, metrics, and baselines, then report held-out test performance and generation efficiency. We further analyze the evolution trajectory on Singapore and illustrate how the evolved heuristic system produces interpretable trip chains and captures empirical mobility patterns.

\subsection{Experimental Setup}
\label{subsec:setup}
\textbf{Datasets.} 
We evaluate \emph{MobEvolve} on two real-world daily mobility generation benchmarks. The \emph{Singapore} dataset is an activity-only task: given individual features, the generator produces a 24-slot trip chain over \emph{Home}, \emph{Work}, \emph{School}, \emph{Others}, and \emph{Travel}. The \emph{Montreal} dataset is a more challenging joint activity-location generation task, where each hourly slot requires both an activity state and a spatial location. For both tasks, system evolving is performed on the validation split, and the selected generator is evaluated on the held-out test split (\emph{i.e.}, $\mathcal{D}_{\mathrm{test}}$). Details are provided in Appendix~\ref{appx:dataset}.

\textbf{Metrics.} 
We use three complementary groups of lower-is-better metrics to evaluate generation quality. The \emph{Individual} score measures whether each generated trip chain recovers the individual's activity or activity-location state at each hourly time slot. The \emph{Distribution} score measures whether the generated population matches empirical mobility distributions. The \emph{Behavior} evaluates whether generated trip chains pass task-specific plausibility tests beyond direct state recovery, such as temporal activity patterns in Singapore and activity-location and land-use consistency in Montreal. The overall score is the sum of the three groups. Metric definitions are provided in Appendix~\ref{appx:metrics}.

\textbf{Baselines.} Our baselines are categorized into three groups: 
(1) \textbf{Traditional mobility models}, including Multinomial Logit (MNL)~\cite{kwak2002multinomial}, Markov~\cite{norris1998markov}, and Gaussian Copula~\cite{SDV}; 
(2) \textbf{Deep Generative Models}, which encompass behavior-choice methods Deep Behavior Choice (DBC)~\cite{wu2024personalized}, Adversarial Inverse Reinforcement Learning (AIRL)~\cite{liang2026analyzing}, alongside deep tabular generative models such as TVAE~\cite{tazwar2024tab}, CopulaGAN~\cite{d2017conscientious}, and CTGAN~\cite{xu2019modeling}; 
(3) \textbf{LLM-based methods}, represented by LLMob~\cite{wang2024large} in both zero-shot and few-shot configurations. Detailed descriptions and configurations for each method are provided in Appendix \ref{appx:baselines}. We perform all model training and inference on a single 32GB V100 GPU, utilizing the PyTorch and the Synthetic Data Vault (SDV\footnote{\url{https://docs.sdv.dev/sdv}}) frameworks. The coding agent employed is Codex, paired with GPT-5.5 w/ High reasoning efforts. To ensure a fair comparison, all methods utilize the exact same dataset splits and metric calculation protocols.

\vspace{-2pt}
\subsection{Performance and Evolution Analysis}
\vspace{-2pt}

\paragraph{Performance on $\mathcal{D}_{\mathrm{test}}$.}
Table~\ref{tab:main-results} shows that MobEvolve achieves the best overall score on both benchmarks. On Singapore, MobEvolve reduces the overall error by 34.1\% relative to the strongest baseline, with the largest gains coming from Distribution and Behavior. This suggests that the evolved heuristics improve population-level calibration and trip-chain plausibility, not just individual hourly-state accuracy. On Montreal, MobEvolve also obtains the best overall score, improving over the strongest baseline by 10.8\% in the more challenging joint activity-location generation setting. The gains are consistent across Individual, Distribution, and Behavior, showing that MobEvolve can improve activity-location recovery while preserving population alignment and behavior consistency. In contrast, LLM-based baselines remain weak on Distribution, and deep models often suffer from high Behavior errors, suggesting that direct generation and black-box distribution learning do not fully satisfy the requirements of mobility generation.

\paragraph{Evolution Trajectory.}

\begin{figure}[h!]
\centering
\includegraphics[width=\linewidth]{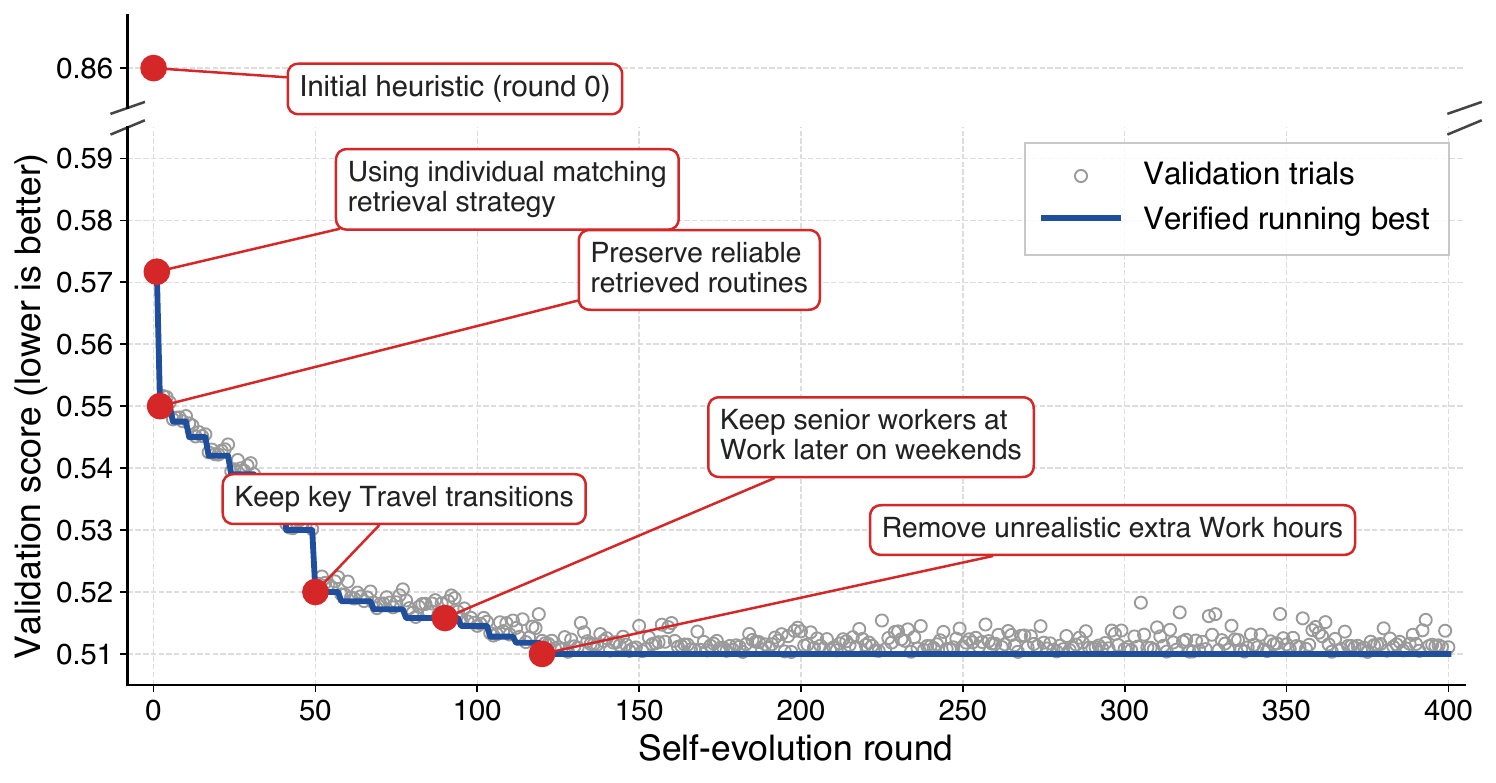}
\vspace{-22pt}
\caption{Self-evolution trajectory on the Singapore validation set. Gray points represent attempted heuristic updates, while the solid line tracks the running best used for selection. Red markers highlight accepted updates that yield major performance improvements.}
\vspace{-15pt}
\label{fig:sg-evolution}
\end{figure}

Figure~\ref{fig:sg-evolution} shows how these gains emerge through self-evolution. Gray points are attempted heuristic updates, while the running-best curve records the MobEvolve selected after verification. The validation score decreases from 0.8621 to 0.5118, a \textbf{40.6\%} relative reduction, but the curve is not a smooth optimization trajectory. Instead, improvements occur when the agent identifies a concrete behavioral failure and proposes a targeted heuristic revision. \textbf{Early} accepted updates make mobility adaptation more conservative and sparse, reducing unnecessary changes to strong retrieved patterns while still repairing low-support hourly choices. This lowers the Behavior score from 0.0983 to 0.0824 in Round 002 and further to 0.0692 after Round 003. Later updates address more specific mobility failures, such as preserving sparse Travel anchors for afternoon distributional calibration, repairing senior weekend work tails, and reducing Work leakage for specific population segments. 
The \textbf{intermediate} phase reveals how self-evolution translates diagnosed failures into targeted heuristic updates. Runs 013-035 address a repeated failure mode in behavioral refinement: long \emph{Travel} sequences were collapsed too aggressively, leading to under-predicted afternoon travel and over-predicted \emph{Work}/\emph{Others} activities. Accepted updates preserve sparse interior \emph{Travel} anchors at deterministic positions, improving the Distribution score from 0.0744 in Run 012 to 0.0595 in Run 036 without changing the upstream mobility retrieval process. \textbf{Later updates} focus on finer population-segment refinements, such as repairing weekend work tails for seniors and suppressing work leakage for non-senior weekend individuals.
The trajectory therefore serves as an audit trail: each performance jump corresponds to an interpretable mobility failure, a heuristic update, and an empirical outcome.

\begin{table}[htbp]
\centering
\scriptsize
\definecolor{ourblue}{HTML}{F0F5FF}
\definecolor{groupgray}{HTML}{F9F9F9}
\caption{Singapore test-generation latency and quality. Runtime is measured as the average milliseconds (ms) per sample. Lower is better ($\downarrow$). Best results are \textbf{bolded}.}
\label{tab:sg-speed}
\resizebox{\columnwidth}{!}{%
\begin{tabular}{lcc}
\toprule
\textbf{Method} & \textbf{Score $\downarrow$} & \textbf{ms/sample $\downarrow$} \\
\midrule
\rowcolor{groupgray}
\multicolumn{3}{l}{\emph{\textbf{$\blacktriangledown$ Traditional Mobility Models}}} \\
MNL & 1.6880 & 3.05 \\
Markov & 3.1443 & \textbf{0.44} \\
Gaussian Copula & 3.4071 & 75.00 \\
\midrule
\rowcolor{groupgray}
\multicolumn{3}{l}{\emph{\textbf{$\blacktriangledown$ Deep Generative Models}}} \\
Deep Behavior Choice & 0.7983 & 42.54 \\
AIRL Behavior Choice & 0.7854 & 41.25 \\
TVAE & 3.1466 & 180.00 \\
CopulaGAN & 3.8276 & 220.00 \\
CTGAN & 4.1770 & 240.00 \\
\midrule
\rowcolor{groupgray}
\multicolumn{3}{l}{\emph{\textbf{$\blacktriangledown$ LLM-based Mobility Generation}}} \\
LLMob Few-shot & 1.4958 & 3200.00 \\
LLMob Zero-shot & 9.1559 & 2200.00 \\
\midrule
\rowcolor{ourblue}
\textbf{MobEvolve (ours)} & \textbf{0.5176} & 7.65 \\
\bottomrule
\end{tabular}
}
\end{table}

\paragraph{Generation Efficiency.}
Table~\ref{tab:sg-speed} reports the average generation latency per sample on the Singapore test set. The results demonstrate that MobEvolve offers a significant computational advantage. While Deep Generative Models and LLM-based methods require tens to thousands of milliseconds to generate a single trajectory, MobEvolve achieves a latency of only \textbf{7.65 ms}. This performance is in the same order of magnitude as lightweight traditional models like MNL (3.05 ms). Thus, MobEvolve effectively breaks the trade-off between accuracy and speed: it attains the highest generation quality while maintaining the high throughput necessary for large-scale urban simulations.

\begin{table}[h]
\centering
\small
\caption{Ablation study on the Singapore dataset. Generators are selected by validation overall score, and held-out test scores are reported. Lower is better ($\downarrow$).}
\label{tab:ablations}
\resizebox{\linewidth}{!}{
\begin{tabular}{lrr}
\toprule
\textbf{Setting} & \textbf{Val} $\downarrow$ & \textbf{Test} $\downarrow$ \\
\midrule
Designed initial heuristics & 0.8621 & 0.8539 \\
Unstructured init. & 7.8753 & 7.0256 \\
\midrule
MobEvolve w/ unstructured init. & 0.8753 & 1.0256 \\
MobEvolve (Seed 2026) & 0.5118 & 0.5176 \\
MobEvolve (Seed 2012) & 0.5171 & 0.5127 \\
\bottomrule
\end{tabular}
} 
\end{table}

\paragraph{Randomness and Initialization Ablation}

Table~\ref{tab:ablations} evaluates the effect of evolution randomness and initialization quality on the Singapore dataset. Changing the random seed from 2026 to 2012 leads to similar validation trajectories and comparable test scores, demonstrating that MobEvolve is not driven by a single ``lucky'' evolution path. In contrast, the unstructured initialization starts from a monolithic generator without the four mobility heuristic components. It exhibits exceptionally poor initial performance, and although the evolution process yields substantial improvements, it ultimately converges to a significantly worse solution. This highlights the importance of the behaviorally grounded initial heuristic system: it provides the agent with meaningful mobility decisions to revise during self-evolution. Details of the unstructured initialization are provided in Appendix~\ref{appx:unstructured-initialization}.

\subsection{Evolved Mobility Heuristics}
\label{subsec:case-study}

\definecolor{actualcol}{HTML}{4B5563} 
\definecolor{dbccol}{HTML}{9A5B5B}    
\definecolor{mobcol}{HTML}{1F6F73}    

\providecommand{\mobseg}[2]{%
\begin{array}{@{}c@{}}
\text{\emph{#1}}\\[1.5pt]
\text{\scriptsize #2}
\end{array}}
\providecommand{\chainarrow}{\;\rightarrow\;}

MobEvolve provides interpretability at two levels: how an individual trip chain is generated, and which mobility regularities are uncovered through self-evolution. We analyze both levels by comparing representative outputs with Deep Behavior Choice (DBC)~\cite{wu2024personalized} and by inspecting accepted heuristic updates. 
In the following trip-chain examples, rows are ordered and colored consistently:
\emph{\textcolor{actualcol}{GT}} denotes the ground-truth trip chain from the target data, \emph{\textcolor{dbccol}{DBC}} denotes the Deep Behavior
Choice baseline, and \emph{\textcolor{mobcol}{ME}} denotes MobEvolve.

\paragraph{Tracing Generated Trip Chains.}
At the individual level, MobEvolve exposes the internal decision logic behind each generated trip chain. Each output can be traced through travel participation, mobility retrieval, mobility adaptation, and behavioral refinement, allowing mismatches to be localized to specific mobility decisions.

For a 55-year-old worker, the actual Singapore trip chain follows a workday structure:
\vspace{-5pt}
\[
\small
\setlength{\arraycolsep}{2pt}
\renewcommand{\arraystretch}{1.25}
\begin{array}{@{}r@{\quad}l@{}}
\textsc{\textcolor{actualcol}{GT}} &
{\color{actualcol}
\mobseg{Home}{00--05}
\chainarrow
\mobseg{Travel}{06}
\chainarrow
\mobseg{Work}{07--17}
\chainarrow
\mobseg{Home}{18--23}}
\\
\textsc{\textcolor{dbccol}{DBC}} &
{\color{dbccol}
\mobseg{Work}{00--07}
\chainarrow
\mobseg{Travel}{08--09}
\chainarrow
\mobseg{Others}{10--18}
\chainarrow
\mobseg{Travel}{19--20}
\chainarrow
\mobseg{Home}{21--23}}
\\
\textsc{\textcolor{mobcol}{ME}} &
{\color{mobcol}
\mobseg{Home}{00--05}
\chainarrow
\mobseg{Travel}{06--07}
\chainarrow
\mobseg{Work}{08--17}
\chainarrow
\mobseg{Home}{18--23}}
\end{array}
\]
\vspace{-2pt}
The comparison highlights two different error modes. DBC breaks the daily structure by placing \emph{Work} at midnight and replacing the main work block with a long \emph{Others} segment. MobEvolve preserves the workday structure and only shifts the commute-to-work transition by one hourly slot. This localized mismatch is consistent with the generator's decision structure: the retrieved daily routine and refinement preserve the overall activity chain, while the remaining error lies in temporal adaptation around the commute boundary.

For a Montreal student case, the task additionally evaluates whether activity states are paired with plausible locations. The sketch below shows the activity-time structure:
\vspace{-5pt}
\[
\small
\setlength{\arraycolsep}{2pt}
\renewcommand{\arraystretch}{1.25}
\begin{array}{@{}r@{\quad}l@{}}
\textsc{\textcolor{actualcol}{GT}} &
{\color{actualcol}
\mobseg{Home}{00--06}
\chainarrow
\mobseg{Travel}{07}
\chainarrow
\mobseg{School}{08--15}
\chainarrow
\mobseg{Travel}{16}
\chainarrow
\mobseg{Home}{17--23}}
\\
\textsc{\textcolor{dbccol}{DBC}} &
{\color{dbccol}
\mobseg{Travel}{00--06}
\chainarrow
\mobseg{School}{07--23}}
\\
\textsc{\textcolor{mobcol}{ME}} &
{\color{mobcol}
\mobseg{Home}{00--06}
\chainarrow
\mobseg{Travel}{07}
\chainarrow
\mobseg{School}{08--15}
\chainarrow
\mobseg{Travel}{16}
\chainarrow
\mobseg{Home}{17--23}}
\end{array}
\]
\vspace{-2pt}
DBC fails at both levels: it misses the return-home structure and keeps the student in an incorrect school region until the end of the day, matching only 3/24 joint activity-location slots. MobEvolve preserves the school-day loop and matches 22/24 joint slots, with the remaining error localized to destination choice within the school block. This case shows that MobEvolve preserves the behavioral structure of the trip chain, while its residual error can be traced to a specific location decision.

\paragraph{Interpreting Evolved Mobility Heuristics.}

On Singapore, evolution reveals daily activity rhythms, accepted updates mainly refine when activities appear and how long they persist. Early updates preserve rare but meaningful events, such as late-night \emph{Travel} and short work tails. Later updates target specific population groups: homemakers avoid worker-like commute patterns, senior weekend workers require different work timing, and late \emph{Work} is preserved only after a sustained \emph{Work} block. These updates show that the Singapore generator learns group-conditioned rhythms, where \emph{Home} anchors most of the day, \emph{Travel} marks sparse transitions, and \emph{Work} appears only when supported by the person's role and surrounding schedule.

On Montreal, evolution is primarily spatial. Early updates repair coarse
trip-chain structure, including telework behavior and student midday home
periods. Later updates focus on regional activity-location allocation, adjusting
\emph{Work} and \emph{School} destinations according to home region, population
group, weekday context, and activity type. These changes reduce over-attraction
to globally frequent destinations and improve neighborhood-specific destination
choices.

\section{Conclusion}

In this work, we presented \textbf{MobEvolve}, the first agentic self-evolving heuristic system for interpretable human mobility generation. MobEvolve addresses the central challenge of synthesizing trip chains that are both behaviorally plausible at the individual level and distributionally aligned at the population level.  MobEvolve starts from behaviorally grounded mobility heuristics and lets an LLM agent evolve their internal generation logic through rollout, distributional-behavioral attribution, heuristic evolution, and memory evolution. Experiments on Singapore and Montreal benchmarks show that MobEvolve achieves the best overall scores while producing interpretable generation processes and evolution traces that uncover mobility regularities. These results suggest that self-evolving mobility heuristics can provide a scalable and auditable path toward adaptive mobility generation, and may also serve as a tool for analyzing empirical mobility behavior.

\clearpage
\section*{Limitations}

MobEvolve improves mobility generation by evolving interpretable mobility heuristics, but it still depends on the quality and coverage of the empirical observations used for reference, evaluation, and evolution. If the reference data under-represents certain population segments, rare travel patterns, or spatial contexts, the evolved heuristics may inherit these gaps. The current experiments focus on daily trip-chain generation in two regional settings, so further evaluation is needed to assess transfer to other cities, finer temporal resolutions, longer horizons, and richer mobility constraints. In addition, although evolution memory makes the update process auditable, the LLM agent may still propose ineffective or overly specific heuristic revisions; MobEvolve therefore requires fixed evaluators, validation splits, and verification procedures to reduce overfitting to evolution feedback.

\section*{Ethical Considerations}

Human mobility data can reveal sensitive information about individuals, households, and communities. MobEvolve is designed for population-level mobility synthesis and evaluation, not for identifying or reconstructing real individuals. Nevertheless, generated trip chains may reflect biases, omissions, or imbalances present in the empirical data and may affect downstream planning analyses if used without careful validation. Applications should therefore use appropriate privacy protection, avoid releasing records that can be linked to real persons, and evaluate performance across population segments. Because MobEvolve produces auditable heuristic updates, its outputs and evolution traces should be inspected before deployment in policy or planning settings, especially when decisions may affect access, equity, or resource allocation.

\bibliography{arxiv}

\clearpage
\appendix

\onecolumn

\clearpage
\phantomsection
\label{appx:contents}

{\huge\textbf{Appendix}}
\vspace{8pt}

\textbf{Table of Contents}\\
\noindent\makebox[\textwidth]{\rule{\textwidth}{0.4pt}}

\textbf{\hyperref[appx:implementation]{A Implementation Details of MobEvolve}} \dotfill \pageref{appx:implementation}

\hspace{0.5cm} \hyperref[appx:heuristic-generator]{A.1 Heuristic Generator} \dotfill \pageref{appx:heuristic-generator}

\hspace{0.5cm} \hyperref[appx:evolution-protocol]{A.2 Evolution Protocol} \dotfill \pageref{appx:evolution-protocol}

\hspace{0.5cm} \hyperref[appx:evolution-context]{A.3 Evolution Context} \dotfill \pageref{appx:evolution-context}

\hspace{0.5cm} \hyperref[appx:system-organization]{A.4 System Organization} \dotfill \pageref{appx:system-organization}

\vspace{3pt}

\textbf{\hyperref[appx:dataset]{B Dataset Details}} \dotfill \pageref{appx:dataset}

\vspace{3pt}

\textbf{\hyperref[appx:metrics]{C Metric Details}} \dotfill \pageref{appx:metrics}

\hspace{0.5cm} \hyperref[appx:metrics-singapore]{C.1 Singapore Activity-Diary Metrics} \dotfill \pageref{appx:metrics-singapore}

\hspace{0.5cm} \hyperref[appx:metrics-montreal]{C.2 Montreal Activity-Location Metrics} \dotfill \pageref{appx:metrics-montreal}

\vspace{3pt}

\textbf{\hyperref[appx:baselines]{D Baseline Details}} \dotfill \pageref{appx:baselines}

\textbf{\hyperref[appx:unstructured-initialization]{E Unstructured Initialization}} \dotfill \pageref{appx:unstructured-initialization}

\noindent\makebox[\textwidth]{\rule{\textwidth}{0.4pt}}

\vspace{5pt}

\section{Implementation Details of MobEvolve}
\label{appx:implementation}

This appendix describes how MobEvolve is implemented in our experiments. The
implementation follows the four-stage procedure in the main text:
\emph{Mobility Rollout}, \emph{Distributional-Behavioral Attribution},
\emph{Heuristic Evolution}, and \emph{Memory Evolution}. The data, evaluator,
validity checks, and evolution protocol are fixed across rounds; only the
internal heuristic logic of the generator is modified during self-evolution.

\subsection{Heuristic Generator}
\label{appx:heuristic-generator}

The generator is implemented as a modular Python system under \texttt{hs/}. Each
call to the generator takes a persona \(x_i\) and returns a 24-slot
activity-location diary
\(\hat{y}_i=\{(\hat{a}_{i,t},\hat{l}_{i,t})\}_{t=1}^{24}\). The activity
vocabulary is
\[
\{\texttt{Home},\texttt{Work},\texttt{School},\texttt{Others},\texttt{Travel}\},
\]
and each location is either a selected SP23 region or the special
\texttt{Travel} location. The implemented pipeline follows
Eq.~\ref{eq:mobility_heuristic_factorization}: it first samples travel
participation, retrieves an empirical routine, adapts the activity-location
sequence, and finally applies behavioral refinement to enforce consistency.

Before each evolution round, MobEvolve rebuilds the train-derived heuristic
state from the reference data \(\mathcal{D}\). This state contains empirical
routine templates, segment-level participation rates, activity marginals,
transition statistics, location marginals, destination-choice evidence, region
land-use features, and spatial centroids. These artifacts provide empirical
evidence for the heuristic components rather than learned neural parameters.

\subsection{Evolution Protocol}
\label{appx:evolution-protocol}

\paragraph{Mobility Rollout.}
At round \(k\), the current generator \(G^{(k)}\) is applied to every persona in
the evolution population \(X_{\mathrm{tar}}\). The resulting
\(\hat{Y}^{(k)}\) is written as a prediction trace and becomes the object of
evaluation and attribution. The rollout uses a fixed random seed so that changes
across rounds can be attributed to heuristic updates rather than uncontrolled
sampling variation.

\paragraph{Distributional-Behavioral Attribution.}
After rollout, MobEvolve compares \(\hat{Y}^{(k)}\) with
\(\mathcal{D}_{\mathrm{tar}}\). The evaluator computes activity, location, and
joint activity-location metrics, including marginal distribution alignment,
transition alignment, paired slot accuracy, non-home recall, segment-level
performance, joint accuracy, and invalid location rate. These evaluator outputs
are compiled into an \texttt{attribution\_record/}, which localizes failures by
time slot, activity-location confusion, persona segment, and representative
individual generated trip chains. The record connects scalar evaluation signals
to behavioral evidence that can guide the next heuristic update.

\paragraph{Heuristic Evolution.}
Given the current generator, the attribution record, and the accumulated
\texttt{trial\_memory/}, the agent proposes one targeted update to the heuristic
system. The editable scope is restricted to the generator modules under
\texttt{hs/}; the data, evaluator, regression checks, and evolution orchestrator
remain frozen. Each round performs a single candidate update and a single
verification run. The candidate is promoted only if it passes all validity checks
and satisfies the empirical acceptance rule relative to the latest accepted
parent generator. Otherwise, the parent generator is carried forward.

\paragraph{Memory Evolution.}
After verification, MobEvolve appends the completed trial to
\texttt{trial\_memory/}. Each trial records the targeted failure mode, the
agent's hypothesis, the concrete heuristic intervention, the empirical outcome,
the accept/reject decision, and the consolidated lesson. Accepted trials provide
positive evidence about effective mobility heuristics, while rejected trials
preserve negative evidence about unsupported changes. This makes later evolution
rounds cumulative rather than independent search attempts.

\subsection{Evolution Context}
\label{appx:evolution-context}

MobEvolve maintains an evolution context that is updated after each heuristic
evolution trial and made available to the agent in later rounds. This context is
not a learned memory module or a separate retrieval model. Instead, it is a
persistent experimental record that allows the agent to relate current
behavioral failures to prior heuristic interventions, empirical outcomes, and
task constraints.

\paragraph{\texttt{attribution\_record/}.}
This record describes the current behavioral misalignment identified by
Distributional-Behavioral Attribution. It links aggregate evaluation signals to
localized evidence, such as affected time periods, population segments,
activity-location confusions, and representative generated trip chains. Its role
is to turn a scalar performance gap into an interpretable failure mode that can
guide heuristic revision.

\paragraph{\texttt{trial\_memory/}.}
This record stores prior Heuristic Evolution trials. Each trial is summarized by
the proposed intervention, the agent's hypothesis, the measured outcome, and the
lesson inferred from the result. Accepted trials provide positive evidence about
heuristics that improved behavioral alignment, while rejected trials provide
negative evidence about changes that failed empirical verification. This memory
helps the agent avoid repeatedly exploring ineffective updates.

\paragraph{\texttt{verification\_record/}.}
This record stores the empirical and structural verification outcome for each
candidate generator. It includes the relevant mobility metrics, validity checks,
comparison to the parent generator, and the final accept/reject decision. Its
purpose is to ensure that heuristic evolution is driven by verified behavioral
improvement rather than by unvalidated or structurally invalid changes.

\paragraph{Persistent artifacts.}
In the implementation, these conceptual records are backed by persistent
artifacts, including the agent's round notes, metric summaries, failure reports,
code diffs, generated trip chains, regression checks, and generator snapshots.
These artifacts are not separate memory mechanisms; they provide the evidence
from which the attribution record, trial memory, and verification record are
constructed. Train-derived index artifacts are treated as part of the heuristic
generator state rather than as evolution memory.

\paragraph{Design rationale.}
The evolution context is designed to preserve the reasoning chain of
self-evolution without reducing it to a sequence of scores or code changes. Each
round contains three questions: what mobility behavior failed, what heuristic
change was attempted, and whether the change was empirically supported. We
therefore organize the context into three conceptual records:
\texttt{attribution\_record/}, \texttt{trial\_memory/}, and
\texttt{verification\_record/}. This separation lets the agent inspect current
failures, reuse prior evidence, and distinguish successful heuristic updates
from rejected ones.

\clearpage

\subsection{System Organization}
\label{appx:system-organization}

\begin{figure*}[h]
\centering
\begin{tcolorbox}[
title=System Architecture of MobEvolve,
width=0.96\textwidth
]
{\scriptsize
\begin{Verbatim}
MobEvolve/
    frozen/
      data/
        train.jsonl                  # reference data D
        val.jsonl                    # target observations D_tar for evolution
        test.jsonl                   # held-out final evaluation
      common.py                      # task constants and validators
      evaluate.py                    # frozen mobility evaluator

    hs/
      generate.py                    # generator entry point
      m0_travelflag.py               # travel participation
      m3_template.py                 # mobility retrieval
      m5_adapt.py                    # activity adaptation
      m5_location_adapt.py           # location adaptation
      m4_repair.py                   # behavioral refinement
      index.py                       # builds train-derived heuristic state
      index_artifacts/
        templates.jsonl
        travel_rate_by_segment.json
        slot_marginals.json
        transitions.json
        location_slot_marginals.json
        destination_choice.json
        location_by_activity.json
        region_features.json
        day_patterns.json

    evolve.py                        # one-round self-evolution orchestrator
    tests/
      regression.py                  # validity and consistency checks

    runs/<run_id>/
      val_predictions.jsonl          # rollout output
      val_metrics.json               # evaluator output
      val_failures.json              # attribution_record/
      codex_notes.md                 # agent hypothesis and intended update
      diff.patch                     # heuristic intervention
      regression.json                # validity checks
      trial_meta.json                # promotion/rejection metadata
      hs_snapshot/                   # generator state for auditability

    trials.jsonl                     # trial_memory/
    NEXT_ROUND.md                    # context passed to the next round
\end{Verbatim}
}
\end{tcolorbox}
\vspace{-8pt}
\caption{Implementation organization of MobEvolve. The frozen components define
the task and evaluator, \texttt{hs/} contains the evolvable heuristic generator,
and each run stores rollout, attribution, intervention, verification, and memory
artifacts.}
\label{fig:system-arch}
\end{figure*}


\section{Dataset Details}
\label{appx:dataset}

Table~\ref{tab:dataset-details} summarizes the two benchmarks. The activity vocabulary is shared across datasets: \emph{Home}, \emph{Work}, \emph{School}, \emph{Others}, and \emph{Travel}. In Montreal, \emph{Home} slots must use the person's home SP23 region, \emph{Travel} slots use a dedicated \emph{Travel} location token.

\begin{table*}[htbp]
\centering
\small
\setlength{\tabcolsep}{4pt}
\renewcommand{\arraystretch}{1.05}
\caption{Dataset statistics. Both datasets represent each diary using 24 activity slots and five activity labels. Trip-maker and multi-trip counts are computed on the full split union. Singapore does not contain explicit destination labels, so the location field is not applicable.}
\label{tab:dataset-details}
\resizebox{\linewidth}{!}{%
\begin{tabular}{lccccccl}
\toprule
\textbf{Dataset} & \textbf{Records} & \textbf{Train} & \textbf{Val} & \textbf{Test} & \textbf{Locations} & \textbf{Trip makers} & \textbf{Multi-trip / multi-Travel structure} \\
\midrule
Singapore & 32,246 & 22,572 & 4,836 & 4,838 & N/A & 26,399 (81.9\%) & 12,507 (38.8\%) w/ $>1$ \emph{Travel} block \\
Montreal & 19,715 & 15,772 & 1,971 & 1,972 & 40 & 19,715 (100.0\%) & 19,342 (98.1\%) w/ $>1$ source trip; mean $=2.84$ \\
\bottomrule
\end{tabular}
}
\end{table*}

\paragraph{Singapore Dataset.}
We utilize the Household Travel Survey data from Singapore. The original survey collected detailed trip information for 35,714 users over a self-reported day between June 25, 2012, and May 30, 2013, including socio-demographic features such as age, income, car ownership, gender, and employment type. To align the raw survey data with our modeling framework, we apply several preprocessing and filtering steps. We filter out records with incomplete or invalid trip sequences and missing socio-demographic information. As summarized in Table~\ref{tab:dataset-details}, this filtering yields a final dataset of 32,246 daily records (diaries) from 26,399 trip makers. Furthermore, we map the original diverse activity purposes into five standardized categories: \emph{Home}, \emph{Work}, \emph{School}, \emph{Others}, and \emph{Travel}. To formulate the activity-travel decision process consistently across benchmarks, we discretize the daily continuous time into 24 hourly intervals (slots). The Singapore dataset does not contain explicit destination labels, so the location field is not applicable.

\paragraph{Montreal Dataset.}
We use a filtered Montréal subset of a 2023 regional household travel survey. Each record corresponds to one trip, and records with the same person-day identifier form an ordered daily trip chain. To ensure high-quality spatial features, the analysis is restricted to 39 Montréal SP23 sectors where the municipal open-data land-use layer covers at least 95\% of the sector area. Complete trip chains are kept only when the home sector and all trip ends fall within these selected sectors. We further filter out chains with missing variables or survey non-response codes in key household and personal attributes. The final dataset contains 55,935 trips from 19,715 complete person-day chains. Notably, 19,342 diaries (98.1\%) have more than one source trip, with a mean chain length of 2.84 trips. To align with our modeling framework, we discretize departure times into 24 hourly slots and map the original survey purposes into the 5 shared activity categories. The spatial geography is defined by 40 discrete locations, comprising the 39 retained SP23 sectors and one dedicated \emph{Travel} token. Furthermore, land-use attributes are attached at the SP23 level. We construct these features from the Montréal open-data \emph{Plan d'urbanisme et de mobilité 2050} land-use layers.\footnote{\url{https://donnees.montreal.ca/dataset/niveaux-intensification-urbaine-densite-affectation-sol-pum-2050}} For each retained sector, we compute the share of covered area assigned to eight composition variables (e.g., economic activities, local green space, mixed use, and residential use), providing rich spatial context for the travel generation task.


\section{Metric Details}
\label{appx:metrics}

This section details the metric inventory behind the three-level score used in Table~\ref{tab:main-results}.  All component scores are lower-is-better.  For both datasets, the final generation-quality score is the sum of an individual-level score, a distribution-level score, and a behavior score:
\begin{equation}
\begin{aligned}
S_{\mathrm{final}}
={}& S_{\mathrm{individual}}
+ S_{\mathrm{distribution}}
+ S_{\mathrm{task}} .
\end{aligned}
\end{equation}

Throughout this section, distributional diagnostics are computed by comparing the empirical distribution induced by the generated diaries with the corresponding empirical distribution in the real split.  We use Jensen--Shannon divergence (JSD) for these diagnostics.  If $P$ is the real empirical distribution, $Q$ is the generated empirical distribution, and $M=(P+Q)/2$, then
\begin{equation}
\begin{aligned}
\mathrm{JSD}(P,Q)
={}& \frac{1}{2}\mathrm{KL}(P\|M)
   + \frac{1}{2}\mathrm{KL}(Q\|M).
\end{aligned}
\end{equation}
Each behavior diagnostic first extracts a discrete value from each relevant diary or slot, forms empirical distributions for real and generated records, and contributes its JSD to the corresponding component score.

\subsection{Singapore Activity-Diary Metrics}
\label{appx:metrics-singapore}

Singapore diaries contain 24 activity slots per person-day.  Each slot takes one of five activity labels:
\begin{equation}
\mathcal{A}_{\mathrm{SG}}
= \{\text{Home},\text{Work},\text{School},\text{Others},\text{Travel}\}.
\end{equation}
The Singapore final score is
\begin{equation}
S_{\mathrm{final}}^{\mathrm{SG}}
= S_{\mathrm{individual}}^{\mathrm{SG}}
+ S_{\mathrm{distribution}}^{\mathrm{SG}}
+ S_{\mathrm{behavior}}^{\mathrm{SG}} .
\end{equation}

\subsubsection{Individual score}

This component measures person-day, slot-aligned activity accuracy.  It uses two paired prediction metrics and converts each higher-is-better metric into an error term:
\begin{equation}
S_{\mathrm{individual}}^{\mathrm{SG}}
=
(1-\mathrm{slot\_accuracy})
+ (1-\mathrm{weighted\_F1}).
\end{equation}

\paragraph{Slot accuracy.}
For each person-day and each of the 24 slots, the generated activity is compared with the real activity at the same slot.  The metric is the fraction of person-slot pairs whose activity label exactly matches.

\paragraph{Weighted F1.}
All person-slot pairs are treated as five-class activity classification examples over Home, Work, School, Others, and Travel.  The weighted F1 score averages class-wise F1 values using class support as weights.

\subsubsection{Distribution score}

This component measures whether the generated diaries match population-level activity marginals, transition structure, and segment-level calibration.  It is a weighted sum of five JSD diagnostics:
\begin{align}
S_{\mathrm{distribution}}^{\mathrm{SG}}
={}&
\mathrm{slot\_marginal\_jsd\_mean}
+2\,\mathrm{slot\_marginal\_jsd\_max} \nonumber\\
&+\mathrm{activity\_share\_jsd}
+1.5\,\mathrm{pair\_transition\_jsd}
+2\,\mathrm{per\_seg\_jsd\_mean\_max}.
\end{align}

\paragraph{Slot-marginal JSD mean.}
For each hour $h\in\{0,\ldots,23\}$, we compute the empirical activity distribution over the five labels in the real records and in the generated records.  The diagnostic is the mean of the 24 hourly JSD values.

\paragraph{Slot-marginal JSD max.}
Using the same 24 hourly activity distributions, this diagnostic takes the maximum hourly JSD.  It penalizes the worst-calibrated hour.

\paragraph{Activity-share JSD.}
We pool all person-slot activity labels across the whole split and compare the overall activity-share distribution between real and generated diaries.

\paragraph{Pair-transition JSD.}
For each diary, adjacent activity pairs are collected from slots $(0,1),(1,2),\ldots,(22,23)$.  The empirical distribution over ordered pairs such as Home$\rightarrow$Travel and Travel$\rightarrow$Work is compared by JSD.

\paragraph{Worst-segment JSD.}
Diaries are grouped by persona segment.  Within each segment, hourly activity distributions are compared between real and generated records, and the segment-level mean JSD is computed.  The diagnostic is the maximum segment-level mean JSD, so it penalizes the worst-calibrated subgroup.

\subsubsection{Behavior score}

This component measures activity-only behavioral plausibility: start/end-of-day states, night behavior, travel volume, first non-home timing, changed transitions, and subgroup-specific behavior.  It is the unweighted sum of 17 JSD diagnostics:
\begin{align}
S_{\mathrm{behavior}}^{\mathrm{SG}}
={}&
d_{\mathrm{first\_slot}}
+ d_{\mathrm{last\_slot}}
+ d_{\mathrm{night\_nonhome}}
+ d_{\mathrm{night\_work}}
+ d_{\mathrm{night\_others}} \nonumber\\
&+ d_{\mathrm{travel\_count}}
+ d_{\mathrm{first\_nonhome}}
+ d_{\mathrm{changed\_pairs}} \nonumber\\
&+ d_{\mathrm{nonwork\_others}}
+ d_{\mathrm{nonwork\_max\_others\_run}}
+ d_{\mathrm{nonwork\_changed\_pairs}} \nonumber\\
&+ d_{\mathrm{student\_school}}
+ d_{\mathrm{student\_travel}}
+ d_{\mathrm{student\_changed\_pairs}} \nonumber\\
&+ d_{\mathrm{employed\_others}}
+ d_{\mathrm{employed\_travel}}
+ d_{\mathrm{employed\_changed\_pairs}} .
\end{align}

\paragraph{Subgroup definitions.}
Student records are records with employment equal to Full time student.  Employed records are records with employment equal to Employed Full-time or Employed Part-time.  Non-work records are records that are neither student nor employed under the above definitions.

\paragraph{First-slot activity distribution.}
For each diary, extract the activity at slot 0.  Compare the distribution over first-slot activity labels.

\paragraph{Last-slot activity distribution.}
For each diary, extract the activity at slot 23.  Compare the distribution over last-slot activity labels.

\paragraph{Night non-home count distribution.}
Define night slots as slots 0--5 and 22--23.  For each diary, count how many night slots have activity different from Home.  Compare the count distribution.

\paragraph{Night Work count distribution.}
Over the same night slots, count the number of Work slots in each diary.  Compare the count distribution.

\paragraph{Night Others count distribution.}
Over the same night slots, count the number of Others slots in each diary.  Compare the count distribution.

\paragraph{Travel-slot count distribution.}
For each diary, count the total number of Travel slots across the 24-hour day.  Compare the count distribution.

\paragraph{First non-home hour distribution.}
For each diary, record the first slot whose activity is not Home.  If the diary is all Home, assign value 24.  Compare the resulting distribution over first non-home times.

\paragraph{Changed-transition pair distribution.}
For each diary, collect adjacent activity pairs only when the activity changes, i.e., pairs $(a_h,a_{h+1})$ with $a_h\neq a_{h+1}$.  Compare the empirical distribution over changed transition pairs.

\paragraph{Non-work Others count distribution.}
For non-work records only, count the number of Others slots in each diary.  Compare the count distribution.

\paragraph{Non-work maximum Others-run distribution.}
For non-work records only, compute the longest contiguous run length of Others in each diary.  Compare the maximum-run distribution.

\paragraph{Non-work changed-transition pair distribution.}
For non-work records only, collect changed adjacent activity pairs.  Compare the subgroup-specific changed-transition distribution.

\paragraph{Student School count distribution.}
For student records only, count the number of School slots in each diary.  Compare the count distribution.

\paragraph{Student Travel count distribution.}
For student records only, count the number of Travel slots in each diary.  Compare the count distribution.

\paragraph{Student changed-transition pair distribution.}
For student records only, collect changed adjacent activity pairs.  Compare the subgroup-specific changed-transition distribution.

\paragraph{Employed Others count distribution.}
For employed records only, count the number of Others slots in each diary.  Compare the count distribution.

\paragraph{Employed Travel count distribution.}
For employed records only, count the number of Travel slots in each diary.  Compare the count distribution.

\paragraph{Employed changed-transition pair distribution.}
For employed records only, collect changed adjacent activity pairs.  Compare the subgroup-specific changed-transition distribution.

\subsection{Montreal Activity-Location Metrics}
\label{appx:metrics-montreal}

Montreal diaries contain both a 24-slot activity sequence and a 24-slot location sequence.  The evaluation therefore measures not only whether the activity is correct, but also whether the generated activity-location pair is spatially and behaviorally plausible.  The Montreal final score is
\begin{equation}
S_{\mathrm{final}}^{\mathrm{MTL}}
= S_{\mathrm{individual}}^{\mathrm{MTL}}
+ S_{\mathrm{distribution}}^{\mathrm{MTL}}
+ S_{\mathrm{behavior}}^{\mathrm{MTL}} .
\end{equation}

\subsubsection{Individual score}

This component measures paired activity-location correctness at the person-day and slot level.  Higher-is-better paired metrics are converted into error terms, and invalid locations are penalized directly:
\begin{equation}
\resizebox{\linewidth}{!}{$
\begin{aligned}
    S_{\mathrm{individual}}^{\mathrm{MTL}}
    ={}&
    (1-\mathrm{slot\_accuracy})
    +(1-\mathrm{joint\_slot\_accuracy}) \\
    &+(1-\mathrm{joint\_weighted\_F1})
    +(1-\mathrm{per\_seg\_joint\_acc\_min})
    +5\,\mathrm{invalid\_location\_rate}.
\end{aligned}
$}
\end{equation} 
\paragraph{Slot accuracy.}
For each person-day and each slot, compare the generated activity with the real activity at the same slot.  Location is not used in this diagnostic.

\paragraph{Joint-slot accuracy.}
Each slot is represented as a joint label activity@location.  A prediction is correct only if both the activity and the location match the real slot.

\paragraph{Joint weighted F1.}
All person-slot pairs are treated as multi-class examples over joint activity@location labels.  Weighted F1 is computed over these joint labels, using class support as weights.

\paragraph{Worst-segment joint accuracy.}
Records are grouped by persona segment.  Within each segment, joint-slot accuracy is computed over activity@location labels.  The diagnostic is the minimum segment-level joint accuracy, converted to error as $1-\mathrm{per\_seg\_joint\_acc\_min}$.

\paragraph{Invalid-location rate.}
This is the fraction of generated slots that violate the hard location-consistency constraints.  It is multiplied by 5 in the individual score to penalize illegal activity-location combinations strongly.

\subsubsection{Distribution score}

This component measures whether generated records match activity marginals, location marginals, segment-level activity distributions, and joint activity-location shares.  It uses eight JSD diagnostics:
\begin{equation}
\resizebox{\linewidth}{!}{$
\begin{aligned}
S_{\mathrm{distribution}}^{\mathrm{MTL}}
={}&
\mathrm{slot\_marginal\_jsd\_mean}
+2\,\mathrm{slot\_marginal\_jsd\_max}
+\mathrm{activity\_share\_jsd} \\
&+2\,\mathrm{per\_seg\_jsd\_mean\_max}
+0.75\,\mathrm{location\_slot\_marginal\_jsd\_mean} \\
&+0.75\,\mathrm{location\_slot\_marginal\_jsd\_max}
+0.75\,\mathrm{location\_share\_jsd}
+0.75\,\mathrm{joint\_share\_jsd}.
\end{aligned}
$}
\end{equation}

\paragraph{Slot-marginal activity JSD mean.}
For each of the 24 slots, compute the empirical activity distribution in real and generated records.  Average the 24 hourly JSD values.

\paragraph{Slot-marginal activity JSD max.}
Using the same hourly activity distributions, take the maximum JSD across the 24 slots.  This captures the worst hour-level activity mismatch.

\paragraph{Activity-share JSD.}
Pool all person-slot activity labels over the split and compare the overall activity-share distribution.

\paragraph{Worst-segment activity JSD.}
Group records by persona segment, compute the mean hourly activity-distribution JSD within each segment, and take the maximum over segments.

\paragraph{Location slot-marginal JSD mean.}
For each slot, compute the empirical distribution over location labels in real and generated records.  Average the 24 hourly location-distribution JSD values.

\paragraph{Location slot-marginal JSD max.}
Using the same hourly location distributions, take the maximum JSD across the 24 slots.

\paragraph{Location-share JSD.}
Pool all person-slot location labels over the split and compare the overall location-share distribution.

\paragraph{Joint-share JSD.}
Pool all person-slot activity@location labels over the split and compare the overall joint activity-location distribution.

\subsubsection{Behavior score}

This component measures whether non-home activities occur in plausible places, whether fixed away locations and travel intensity match the real data, and whether Work/Others locations and land-use semantics are calibrated.  It is the unweighted sum of 10 JSD diagnostics:
\begin{align}
S_{\mathrm{behavior}}^{\mathrm{MTL}}
={}&
d_{\mathrm{home\_region\_nonhome}}
+ d_{\mathrm{other\_unique\_nonhome\_loc}}
+ d_{\mathrm{away\_fixed\_loc\_slots}} \nonumber\\
&+ d_{\mathrm{others\_loc}}
+ d_{\mathrm{nonhome\_landuse}}
+ d_{\mathrm{night\_nonhome}} \nonumber\\
&+ d_{\mathrm{worker\_travel}}
+ d_{\mathrm{joint\_activity\_loc\_share}}
+ d_{\mathrm{work\_loc}}
+ d_{\mathrm{work\_landuse}} .
\end{align}

\paragraph{Behavior diagnostic setup.}
In these diagnostics, a fixed non-home location means a slot whose activity is Work, School, or Others and whose location is neither the person's home SP23 region nor Travel nor an unknown location label.  Employment groups are derived from the Montreal status field: worker records use status values 1 or 2, student records use status value 3, and Other-population records use status value 5.

\paragraph{Home-region non-home activity count distribution.}
For each diary, count slots whose activity is Work, School, or Others but whose location equals the person's home SP23 region.  Compare the distribution of these counts.  This penalizes unrealistic non-home activities that remain in the home region too often or too rarely.

\paragraph{Other-population unique non-home location count distribution.}
For records in the Other-population group, count the number of unique fixed non-home locations visited in a diary.  Compare the distribution of unique-location counts.  This measures whether non-worker, non-student diaries have realistic spatial variety.

\paragraph{Away fixed-location slot count distribution.}
For each diary, count slots whose activity is Work, School, or Others and whose location is a fixed away location rather than home or Travel.  Compare the count distribution.  This captures the overall volume of away-location activity.

\paragraph{Others location distribution.}
Extract all slots whose activity is Others and whose location is a valid non-Travel location.  Compare the empirical distribution over Others locations.  This measures whether discretionary or other-purpose activities occur in realistic places.

\paragraph{Non-home land-use distribution.}
For all fixed non-home Work, School, and Others slots, map each location to its dominant land-use class using the Montreal SP23 land-use table.  Compare the distribution over dominant land-use classes.  This measures whether non-home activities occupy plausible urban land-use contexts.

\paragraph{Night non-home count distribution.}
Define night slots as slots 0--5 and 20--23.  For each diary, count the number of night slots whose activity is not Home.  Compare the count distribution.  This measures nighttime activity plausibility.

\paragraph{Worker Travel count distribution.}
For worker records only, count the number of Travel slots in each diary.  Compare the distribution of worker Travel counts.  This captures whether worker mobility intensity is realistic.

\paragraph{Joint activity-location share distribution.}
Pool all person-slot activity@location labels over the split and compare the overall joint distribution.  This behavior diagnostic complements the distribution-level joint-share JSD by directly entering the behavior score.

\paragraph{Work location distribution.}
Extract all slots whose activity is Work and whose location is a valid non-Travel location.  Compare the empirical distribution over Work locations.  This measures workplace destination calibration.

\paragraph{Work land-use distribution.}
For all valid Work-location slots, map the location to its dominant land-use class and compare the land-use distribution.  This measures whether generated Work activities are placed in plausible land-use contexts.


\section{Baseline Details}
\label{appx:baselines}

\paragraph{Deep Behavior Choice~\cite{wu2024personalized}.}
This neural baseline trains a supervised slot-level behavior-choice model from scratch on the training split. It conditions on persona attributes and temporal context, predicts the 24-slot diary sequence, and is selected according to validation performance before held-out test evaluation. For Montreal, the same protocol is used to predict joint activity-location outputs.

\paragraph{AIRL Behavior Choice~\cite{liang2026analyzing}.}
This baseline uses the same behavior-choice prediction setting as Deep Behavior Choice, but augments supervised training with an adversarial imitation objective. The discriminator provides an additional trajectory-level signal that encourages generated diaries to resemble real person-day records. All model parameters are trained from scratch and selected according to validation performance.

\paragraph{MNL~\cite{kwak2002multinomial}.}
The multinomial-logit baseline is a classical discrete-choice model over slot labels. It uses observed persona and temporal covariates as explanatory variables and predicts the most likely activity label at each slot. For Montreal, the output space is extended to joint activity-location labels. Predictions are made independently across slots.

\paragraph{Markov~\cite{norris1998markov}.}
The Markov baseline estimates empirical transition probabilities from training diaries and generates each diary sequentially. It captures local state persistence and one-step transition structure, but does not explicitly model persona-conditioned long-range schedule constraints.

\paragraph{SDV generators~\cite{patki2016synthetic}.}
TVAE~\cite{tazwar2024tab}, CTGAN~\cite{xu2019modeling}, CopulaGAN~\cite{d2017conscientious}, and Gaussian Copula~\cite{sklar1959fonctions} are single-table synthetic-data baselines. We flatten each 24-slot diary into one tabular row, train each generator on the training split without test-persona conditioning, sample synthetic rows, and decode the sampled rows back into diary sequences.

\paragraph{LLMob~\cite{wang2024large}.}
The zero-shot LLMob baseline receives only the persona and task schema, then generates one 24-slot diary in a single forward pass. The few-shot LLMob baseline additionally receives three randomly selected demonstrations from the same broad occupation group in Singapore or the same status group in Montreal, including both personas and gold diaries. We use the Qwen-flash API\footnote{\url{https://www.alibabacloud.com/help/en/model-studio/models}} as the underlying LLM.


\section{Unstructured Initialization}
\label{appx:unstructured-initialization}

The unstructured initialization is designed as a stress test of the structural prior used by MobEvolve. Unlike the modular mobility decomposition used in our main system, this initialization starts from a single monolithic diary generator. The generator first emits an all-\emph{Home} day and then applies only a small set of coarse rules: full-time students on weekdays receive a fixed \emph{School} block, full-time workers on weekdays receive a fixed \emph{Work} block, part-time and self-employed workers occasionally receive a shorter \emph{Work} block, and a small fallback probability inserts a single \emph{Others} activity.

This initialization deliberately omits several components used by MobEvolve, including an explicit travel-decision module, template retrieval, marginal or Markov sampling, local adaptation, segment abstraction, and schedule repair. As a result, it lacks several basic mobility priors: commute shoulders around work and school episodes, realistic first non-home timing, night-tail behavior, weekday/weekend specialization, long-\emph{Travel} handling, work-from-home behavior, and subgroup-specific schedule repair.

\end{document}